\documentclass[10pt,twocolumn,letterpaper]{article}

\usepackage[final]{cvpr}      
\usepackage{graphicx}
\usepackage{amsmath}
\usepackage{amssymb}
\usepackage{booktabs}
\usepackage[table]{xcolor}
\usepackage{threeparttable}
\usepackage[accsupp]{axessibility} 
\usepackage[pagebackref,breaklinks,colorlinks,citecolor=cvprblue]{hyperref}
\usepackage[capitalize]{cleveref}
\crefname{section}{Sec.}{Secs.}
\Crefname{section}{Section}{Sections}
\Crefname{table}{Table}{Tables}
\crefname{table}{Tab.}{Tabs.}

\usepackage{xcolor}
\usepackage{enumitem}
\usepackage{xspace}
\usepackage{booktabs}
\usepackage{colortbl}
\usepackage{multirow}
\usepackage{makecell}
\usepackage{lipsum} 
\usepackage{gensymb} 

\newcommand{\Tref}[1]{Table~\ref{#1}}

\newcommand{\Eref}[1]{Equation~\eqref{#1}}

\usepackage[labelsep=period]{caption}
\renewcommand{\paragraph}[1]{\vspace{0.2em}\noindent \textbf{#1 \hspace{0.2em}}}

\definecolor{MyDarkRed}{rgb}{0.66, 0.16, 0.16}
\definecolor{MyDarkBlue}{rgb}{0.16, 0.16, 0.66}

\definecolor{cvprblue}{rgb}{0.21,0.49,0.74}

\def\paperID{12302} 
\def\confName{CVPR}
\def\confYear{2024}

\title{Towards Real-World HDR Video Reconstruction: A Large-Scale Benchmark Dataset and A Two-Stage Alignment Network }
\author{Yong Shu, \space Liquan Shen\thanks{Corresponding author. This work was partially supported by China NSFC grant (no.61931022 and no.62271301), Shanghai SSTP grant (22511105200), Shanghai SEALP grant (23XD1401400).}, \space Xiangyu Hu, \space Mengyao Li, \space Zihao Zhou\\
Shanghai University, China\\
{\tt\small $\lbrace$yungsyu,jsslq,arhu314,sdlmy,yi$\_$yuan$\rbrace$@shu.edu.cn}
}

\begin{document}
\maketitle
\begin{abstract}
   As an important and practical way to obtain high dynamic range (HDR) video, HDR video reconstruction from sequences with alternating exposures is still less explored, mainly due to the lack of large-scale real-world datasets. Existing methods are mostly trained on synthetic datasets, which perform poorly in real scenes.
   In this work, to facilitate the development of real-world HDR video reconstruction, we present \textbf{Real-HDRV}, a large-scale real-world benchmark dataset for HDR video reconstruction, featuring various scenes, diverse motion patterns, and high-quality labels. 
   Specifically, our dataset contains 500 LDRs-HDRs video pairs, comprising about 28,000 LDR frames and 4,000 HDR labels, covering daytime, nighttime, indoor, and outdoor scenes. To our best knowledge, our dataset is the largest real-world HDR video reconstruction dataset.
   Correspondingly, we propose an end-to-end network for HDR video reconstruction, where a novel two-stage strategy is designed to perform alignment sequentially.    
   Specifically, the first stage performs global alignment with the adaptively estimated global offsets, reducing the difficulty of subsequent alignment. 
   The second stage implicitly performs local alignment in a coarse-to-fine manner at the feature level using the adaptive separable convolution. 
   Extensive experiments demonstrate that: (1) models trained on our dataset can achieve better performance on real scenes than those trained on synthetic datasets; (2) our method outperforms previous state-of-the-art methods. Our dataset is available at \textcolor{magenta}{https://github.com/yungsyu99/Real-HDRV.}
\end{abstract}    
\section{Introduction}
\label{sec:intro}
\begin{figure}[t] 
	\centering 
	\hspace{0.35cm}
	\includegraphics[width=0.45\textwidth]{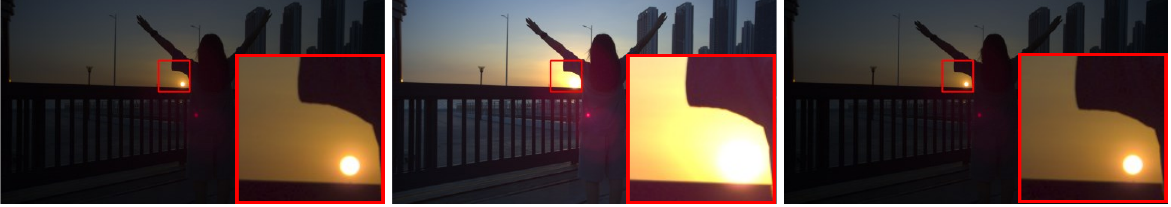}
	\\
	\vspace{-0.22cm}
	\makebox[0.108\textwidth]{\scriptsize \hspace{1.1em} Alternatingly-exposed LDR inputs (\textit{Middle frame} is the reference frame)}
	\\
	\vspace{+0.05cm}
	\hspace{-0.2cm}
	\rotatebox{90}{\scriptsize \hspace{1.6em}\textbf{\makecell{Our\\dataset}} \hspace{1.8em} \makecell{Synthetic\\dataset}}
	\hspace{-0.2cm}
	\includegraphics[width=0.45\textwidth]{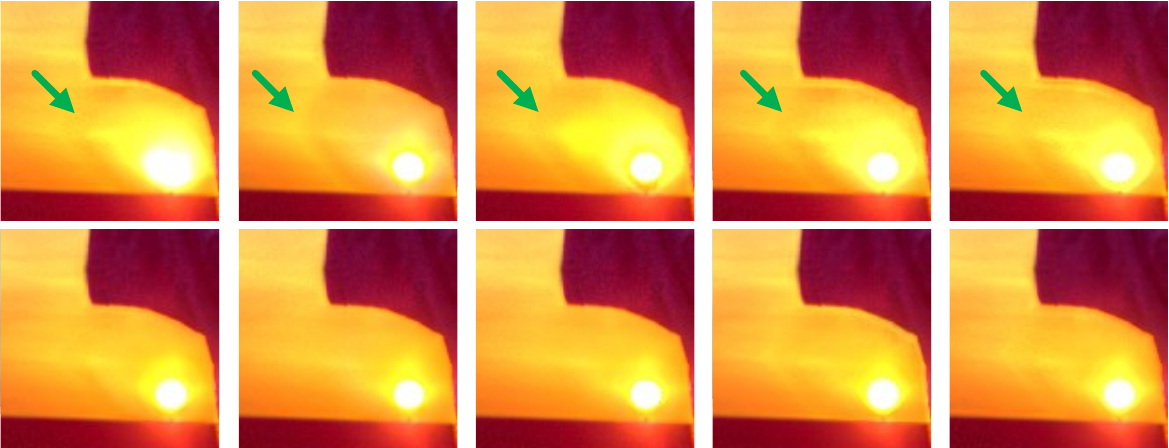}
	\\
	\vspace{-0.17cm}
	\hspace{0.1cm}
	\makebox[0.09\textwidth]{\scriptsize \hspace{0.2cm} AHDRNet\cite{yan2019attention}}
	\makebox[0.09\textwidth]{\scriptsize \hspace{0.1cm} Kalantari19\cite{kalantari2019deep}}
	\makebox[0.08\textwidth]{\scriptsize  \hspace{0.16cm} Chen21\cite{chen2021hdr}}
	\makebox[0.08\textwidth]{\scriptsize \hspace{0.3cm} CA-ViT\cite{liu2022ghost}}
	\makebox[0.09\textwidth]{\scriptsize \hspace{0.34cm} LAN-HDR\cite{chung2023lan}}
	\\
	\caption{Row 1 shows a real-world sample from the Chen21 dataset\cite{chen2021hdr}. Row 2-3 show the HDR frames reconstructed by models trained on the synthetic dataset~\cite{chen2021hdr} and our Real-HDRV, respectively. Obviously, models trained on our dataset are able to recover more and better details of the over-exposed regions. } 
	\label{fig:figure1}
	\vspace{-0.4cm}
\end{figure}
The demands for high dynamic range (HDR) video have drastically increased in recent years since it can bring a better visual experience for users~\cite{kalantari2013patch, gryaditskaya2015motion, barman2021user, xu2024hdrflow}. However, most cameras cannot capture HDR videos directly due to the limitations of sensors. Therefore, some specialized hardware devices \cite{Huggett2009, mcguire2007optical,nayar2000high,Rebecq2021,Han2023} are developed to capture HDR videos. However, these devices are typically bulky and expensive, which are not widely adopted~\cite{chung2023lan}.

In contrast, the computational-based HDR video reconstruction \cite{Kang2003, mangiat2010high} is more practical and affordable for obtaining HDR videos. It captures low dynamic range (LDR) sequences with alternating exposures (\eg, sequences with exposure values of $\lbrace$-3,0,-3,0,...$\rbrace$), which are then used to reconstruct the corresponding HDR video. The common reconstruction pipeline is to align the input frames and then merge the aligned inputs to reconstruct the HDR videos. Before the era of deep learning, some optimization-based reconstruction methods \cite{Kang2003, mangiat2010high, kalantari2013patch} are proposed.
Recently, learning based methods \cite{kalantari2019deep,chen2021hdr,chung2023lan} have shown their effectiveness on HDR video reconstruction, which significantly improve the performance over optimization-based methods.
 
Despite remarkable progress, the development of deep models for HDR video reconstruction is relatively slow, mainly due to the lack of suitable training datasets. The only publicly accessible labeled real-world dataset of HDR video reconstruction \cite{chen2021hdr} is built for evaluating HDR video reconstruction methods. The number of distinct scenes and the motion patterns in their dataset are limited, making it unsuitable for supervised training. Therefore, existing models are still trained on synthetic datasets. However, the synthetic datasets are not well suited for the study of real-world HDR video reconstruction. Models trained on synthetic datasets are hard to generalize to real scenes (see Fig.~\ref{fig:figure1} Row 2) since the synthetic degradations are far different from the real degradations (\eg, the noise in under-exposed areas, the saturation in over-exposed areas). It is highly desired for a large-scale real-world dataset to facilitate the development of real-world HDR video reconstruction.

Besides the datasets, another key issue of HDR video reconstruction lies in the alignment of input frames. 
Previous methods~\cite{Kang2003,kalantari2013patch,kalantari2019deep} usually use optical flow or both optical flow and deformable convolution \cite{chen2021hdr, yue2023hdr} to align the inputs. However, the estimated flows are prone to be inaccurate due to the noise and saturation in alternatingly-exposed inputs, resulting in ghosting artifacts.
Recently, Chung \etal~\cite{chung2023lan} proposed a luminance-based attention module to align the inputs. However, it cannot properly deal with the under-exposed and over-exposed areas of inputs, resulting in unpleasing artifacts. Additionally, the global motion (caused by camera movements) is not properly modeled in most existing methods~\cite{yue2023hdr, chung2023lan}, which further increases the difficulty of alignment, leading to inferior performance.

Based on the above observations, to facilitate the development of real-world HDR video reconstruction, we build a large-scale real-world dataset, named \textbf{Real-HDRV}. 
In order to get LDRs-HDRs video pairs and ensure the quality of HDR labels, we capture the scene in a frame-by-frame manner using a camera with high continuous shooting speed (up to 40 frames/sec). 
Specifically, we carefully select the relatively static scenes and manually create different types of motion between neighboring frames. For each static frame, we capture a multi-exposure image stack (7 differently-exposed LDR images guarantee the quality of HDR labels). The images in each multi-exposure stack are then used to synthesize the corresponding HDR label.
We collected 500 LDRs-HDRs video pairs, comprising about 28,000 LDR frames and 4,000 HDR labels. 
Our Real-HDRV cannot only serve as a benchmark for HDR video reconstruction but also be applied to other HDR tasks (\eg, HDR Deghosting~\cite{yan2019attention}, single-image HDR reconstruction~\cite{Zou_2023_ICCV}).

Correspondingly, we propose an end-to-end network for HDR video reconstruction, in which we design a two-stage strategy to align the inputs sequentially. Specifically, the first stage performs global alignment with the designed global alignment module (GAM), which can effectively handle the global motion and reduce the difficulty of subsequent alignment. The second stage implicitly performs local alignment at the feature level in a coarse-to-fine manner with the designed local alignment module (LAM). The pyramid structure of LAM facilitates the feature alignment under large motion. The adaptive separable convolution~\cite{Niklaus2017} used in LAM enables flexibly integrating the useful information in neighboring frames to compensate for the missing content in the reference frame, which facilitates the feature alignment under noise and saturation. Then, a reconstruction module is applied to reconstruct the HDR video from the aligned features. Our two-stage alignment network can effectively handle complex motion and reconstruct high-quality HDR video. In summary, our contributions are as follows:
\begin{itemize}[itemsep=0pt,parsep=0pt,topsep=2bp]
	\item
	We propose a large real-world HDR video reconstruction dataset, featuring various scenes, diverse motion patterns, and high-quality labels. Our dataset cannot only serve as a benchmark for HDR video reconstruction but also be applied to other HDR imaging tasks.
	\item 
	We propose an end-to-end network for HDR video reconstruction, in which we design a two-stage strategy to perform alignment sequentially. Our network can effectively handle complex motion and achieve high-quality HDR video reconstruction. 
	\item 
	Extensive experiments demonstrate the superiority of our dataset and our method. Our work provides a new platform for researchers to explore real-world HDR video reconstruction techniques.
\end{itemize}

\section{Related Work}


\paragraph{HDR Image Reconstruction} Many methods\cite{Reinhard2002, Rempel2007, Chen2021, Hu2022, Le2023} attempt to perform HDR reconstruction from a single LDR image. However, these methods cannot effectively handle the noise and saturation due to the limited information in a single image. There are methods\cite{sen2012robust, Hu2013} for HDR reconstruction from multi-exposure LDR images. Although these methods work well for static scenes, they generally suffer from ghosting artifacts when tackling dynamic scenes. Therefore, many HDR deghosting methods \cite{gallo2015hdr, yan2019attention, liu2022ghost, CatleyChandar2022, Yan2023} are proposed to alleviate this issue.   

\paragraph{HDR Video Reconstruction Datasets} Kalantari \etal~\cite{kalantari2013patch} captured 5 LDR sequences with two alternating exposures. To quantitatively evaluate HDR video reconstruction methods, Chen \etal~\cite{chen2021hdr} collected 76 dynamic image pairs, 49 static image pairs, and 50 unlabeled LDR sequences with two alternating exposures. However, the number of distinct scenes and the motion patterns in their dataset are limited, making it unsuitable for supervised training. Recently, Yue \etal~\cite{yue2023hdr} collected 85 real-world LDRs-HDRs video pairs using a mobile phone, but, until now, they are not publicly accessible. In addition, they use two images with different exposures to generate an HDR label, which may not cover the full dynamic range of the scene, resulting in limited-quality HDR labels. Due to the lack of publicly accessible large-scale real-world datasets, existing models are still trained on the synthetic dataset~\cite{chen2021hdr}, which hinders the development of real-world HDR video reconstruction.
\renewcommand\arraystretch{1.2}
\begin{table}
	\caption{Comparison between different datasets.}
	\vspace{-0.4cm}
	\label{simple comparison}
	\begin{center}
	\resizebox{\linewidth}{!}{
	\begin{threeparttable}
			\begin{tabular}{c|cccc}
				\toprule
				Dataset&GT&Numbers &Motion patterns&Scenes \\
				\hline
				Chen21 (Dynamic)~\cite{chen2021hdr}&  central frame & 76 & LM&ID \\
				\hline
				Chen21 (Static)~\cite{chen2021hdr}& \makecell{only one \\ static frame} & 49 & Static &ID, IN, OD, ON\\
				\hline
				Ours& per frame &500 &GM, LM, FM&ID, IN, OD, ON \\
				\toprule
		    \end{tabular}
	\end{threeparttable}}
	\end{center}
	\vspace{-0.34cm}
	\scriptsize
	1. Our dataset contains per-frame HDR labels, while the Chen21 dataset only contains the HDR labels for the center frames.\\ 
	2. GM and LM denote global motion (where only the camera is moving) and local motion (where only the foreground is moving), respectively. FM denotes full motion (where both foreground and camera are moving).\\ 
	3. OD, ON, ID and IN denote outdoor daytime, outdoor nighttime, indoor daytime and indoor nighttime, respectively.
	\vspace{-0.3cm}

\end{table}

\paragraph{HDR Video Reconstruction} There are mainly two types of methods to obtain HDR videos: hardware-based methods and computational-based methods. The hardware-based methods~\cite{nayar2000high,mcguire2007optical,Rebecq2021,Han2023} typically rely on specialized hardware systems (\eg, beam splitter), which are typically too expensive to be widely adopted. 

The computational-based methods reconstruct the HDR video from alternatingly-exposed sequences.
Kang \etal~\cite{Kang2003} proposed the first method in this direction, which used optical flow to align the input frames and then merged the aligned frames to generate HDR videos.
Mangiat~\etal~\cite{mangiat2010high} improved \cite{Kang2003} by introducing a block-based motion estimation method with a refinement stage for ghost removal.
Kalantari \etal~\cite{kalantari2013patch} proposed a patch-based method to synthesize the missing exposures at each frame, and these synthesized images are then fused into an HDR frame. 
\\
\indent
Recently, learning based methods have shown their effectiveness on HDR video reconstruction.
Kalantari~\etal~\cite{kalantari2019deep} proposed a flow-based framework, which consists of an optical flow network for alignment and a weight network for merging images.
Chen~\etal~\cite{chen2021hdr} and Yue~\etal~\cite{yue2023hdr} used both optical flow and deformable convolution to perform alignment for reconstructing HDR videos. 
Unfortunately, these methods typically generate ghosting artifacts since the estimated flows are prone to be inaccurate due to the noise and saturation. 
More recently, Chung~\etal~\cite{chung2023lan} proposed a luminance-based alignment network for HDR video reconstruction. However, it cannot properly deal with the under-exposed and over-exposed areas of inputs, resulting in unpleasing artifacts.

\section{Proposed Dataset}
To favor the development of real-world HDR video reconstruction, we construct a large-scale real-world HDR video reconstruction dataset, named Real-HDRV.
Actually, it is extremely challenging to simultaneously capture alternatingly-exposed LDR sequences and the corresponding HDR sequences for dynamic scenes. One may use a beam splitter and two cameras to build a complex optical system to capture two videos with different exposures simultaneously and then generate the HDR video using the captured two videos. However, the amount of light is halved by the beam splitter\cite{Wang2021, froehlich2014creating}, which limits the quality of HDR labels.  
Instead of relying on complex optical systems, to get LDRs-HDRs video pairs and ensure the quality of the HDR labels,  we capture LDRs-HDRs video pairs in a frame-by-frame manner. We manually create motions between neighboring frames and capture a multi-exposure image stack (7 LDR images) for each static frame. \\
\indent
One crucial problem that needs to be addressed is how to ensure that the high-quality HDR label can be obtained after capturing each multi-exposure image stack. The details are as follows:
(1) We use a camera with high continuous shooting speed (up to 40 frames/sec), the Canon R6 Mark2, which enables us to capture a multi-exposure image stack in a very short period of time with one depression of the wireless shutter. 
During the capturing, the camera can almost avoid introducing extra motion, such as camera shake, object movement, etc.
(2) We carefully select the relatively static scenes and manually create different types of motion (\ie, global motion, local motion, and full motion) between neighboring frames to make the motion controllable and diverse. In addition, the camera is mounted on a tripod, and a wireless remote controller is used to avoid introducing extra motion caused by the shutter release. 

Thanks to the high-speed shooting performance of the camera and the careful shooting procedure, we can ensure there is almost no motion between the images in each multi-exposure image stack. Also, each multi-exposure image stack can provide enough images with different exposures (7 images with different exposures guarantee to cover the full dynamic range of a scene). Then, the per-frame high-quality HDR label can be generated from images in each multi-exposure image stack by using the method in \cite{Debevec1997a}, and the LDR images can be arranged in a periodic exposure to generate sequences with alternating exposures\footnote{Since we focus on HDR video reconstruction from sequences with two alternating exposures, we selected LDR frames in a periodic exposure (\ie, $\lbrace$EV-3, EV0, EV-3, EV0, ...$\rbrace$, $\lbrace$EV-2, EV+1, EV-2, EV+1, ...$\rbrace$ or $\lbrace$EV-1, EV+2, EV-1, EV+2, ...$\rbrace$.) to generate the sequences with two alternating exposures. Note that the LDR frames can be selected in different exposure orders for HDR video reconstruction.}. Therefore, we can collect LDRs-HDRs video pairs in a frame-by-frame manner. Specifically, for each frame, we capture a multi-exposure image stack of 7 LDR images spaced by $\pm i$-EV difference where $i\in$ 1, 2, 3  around the reference exposure. Then, we manually create different types of motion between neighboring frames to capture the following image stacks. Finally, all the image stacks are grouped in their temporal order to generate the LDRs-HDRs video pairs. 

\begin{figure*}[!t]
	\centering
	\hspace{0.05cm}
	\includegraphics[width=6.6in]{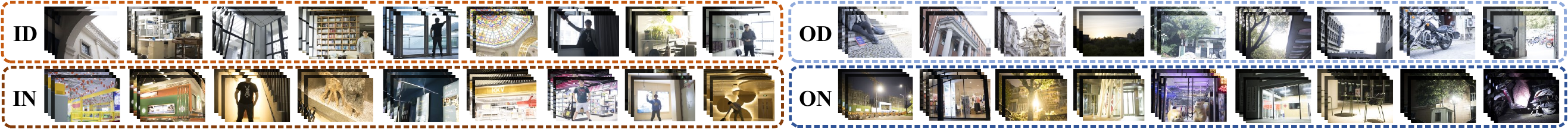}
	\\
	\vspace{-0.17cm}
	\makebox[0.108\textwidth]{\scriptsize (a)}
	\\
	\hspace{0.1cm}
	\includegraphics[width=6.5in]{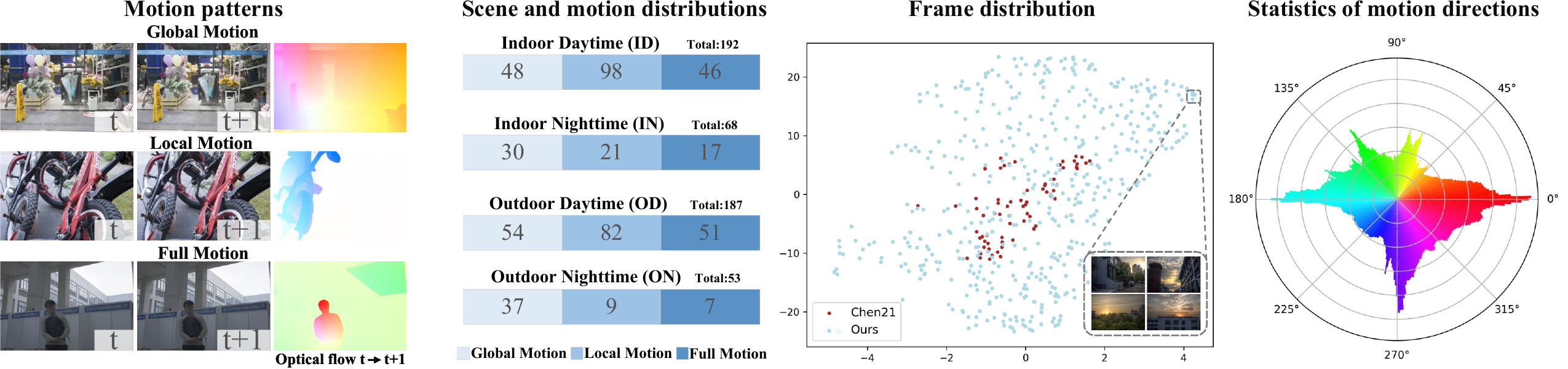}
	\\
	\vspace{-0.18cm}
	\hspace{-0.5cm}
	\makebox[0.2\textwidth]{\scriptsize \hspace{-0.5cm} (b)}
	\makebox[0.2\textwidth]{\scriptsize \hspace{+0.8cm} (c)}
	\makebox[0.2\textwidth]{\scriptsize \hspace{2.1cm} (d)}
	\makebox[0.2\textwidth]{\scriptsize \hspace{2.89cm} (e)}
    \vspace{-0.1cm}
	\caption{(a) Some typical scenes in our dataset, which can be categorized into 4 categories: indoor daytime (ID), indoor nighttime (IN), outdoor daytime (OD), and outdoor nighttime (ON) scenes. (b) Our dataset contains three kinds of motion: global motion (where only the camera is moving), local motion (where only the foreground is moving), and full motion (where both foreground and camera are moving). (c) Scene and motion distributions of our dataset. (d) Diversity comparison: our dataset \textit{vs.} the Chen21 dataset~\cite{chen2021hdr}. (e) Statistics of motion directions in our dataset. We plot a circular histogram, where the color of each bin represents the direction of motion, and the height of the bar represents the proportion of specific directions to all the directions. The per-pixel flow in each frame is computed via RAFT\cite{Teed2020}. } 
	\label{fig:figure2}
	\vspace{-0.3cm}

\end{figure*}

In total, we collected 500 LDRs-HDRs video pairs, each containing 7 to 10 frames. For each frame, 7 differently-exposed LDR images and a high-quality HDR label can be provided. All the images are captured in RAW format with resolution of 6000$\times$4000. We performed the demosaicing, white balancing, color correction, and gamma compression ($\gamma$ = 2.2) to convert the raw data to RGB data. In this work, we rescaled the images to 1500$\times$1000 for training and testing.  Figure~\ref{fig:figure2} shows some typical scenes in our dataset and the statistical indicators of our dataset. In addition, since our dataset provides data in RAW format, the data processing pipeline is highly flexible. Therefore, our dataset can be easily adjusted to make training data for different HDR tasks for future research. 
                 
\renewcommand\arraystretch{1.2}
\begin{table}[!t]
	\caption{Metrics to assess the diversity of different datasets.}
	\vspace{-0.44cm}
	\centering
	\begin{center}
		\label{dataset_metrics}
		\resizebox{\linewidth}{!}{
		\begin{threeparttable}
			\begin{tabular}{c|c}
				\hline
				\multicolumn{2}{c}{\textbf{Metrics on the extent of HDR }}\\
				\hline
				\multirow{1}{*}{\textbf{FHLP}}
				&\hspace{-1cm}Fraction of HighLight Pixel: defined in~\cite{guo2023learning}\\
				\hline
				\multirow{1}{*}{\textbf{EHL}}
				&\hspace{-1cm}Extent of HighLight: defined in~\cite{guo2023learning}\\
				\hline
				\multicolumn{2}{c}{\textbf{Metrics on intra-frame diversity }}\\
				\hline
				\multirow{1}{*}{\textbf{SI}}
				&\hspace{-1cm}Spatial Information: defined in~\cite{SI} \\
				\hline
				\multirow{1}{*}{\textbf{CF}}
				&\hspace{-1cm}ColorFulness: defined in~\cite{hasler2003measuring}\\
				\hline
				\multirow{1}{*}{\textbf{stdL}}
				&\hspace{-1cm}standard deviation of Luminance: defined in~\cite{guo2023learning}\\
				\hline
				\multicolumn{2}{c}{\textbf{Metrics on the overall-style}}\\
				\hline
				\textbf{ALL}&\hspace{-1cm}Average Luminance Level: defined in~\cite{guo2023learning}\\
				\hline
				\multirow{2}{*}{\textbf{DR}}
				&Dynamic Range~\cite{Hu2022}: calculated as the log10 differences between  \\
				&the highest 2$\%$ luminance and the lowest 2$\%$ luminance.\\
				\hline
			\end{tabular}
			\scriptsize
			\vspace{-0.15cm}
			\item 
			Among these aspects, greater extent of HDR represents more probability for the network to learn pixel in advanced HDR volume beyond LDR's capability, higher intra-frame diversity means that the network may learn better generalization capability. We use these metrics to verify the diversity of our dataset
		\end{threeparttable}}
	    \vspace{-0.7cm}
	\end{center}
\end{table}

\paragraph{Analysis of Our Dataset}
To quantitatively evaluate the superiority of our dataset, we analyzed the diversity of the Chen21 dataset\cite{chen2021hdr} and our dataset. Following \cite{Hu2022, guo2023learning}, the 7 metrics defined in \Tref{dataset_metrics} are utilized to assess the diversity of different datasets from 3 aspects, including the extent of HDR, the intra-frame diversity and the overall style of HDR. 
For each HDR label, 7 different metrics are calculated according to \Tref{dataset_metrics}. Then, we use the t-SNE \cite{vd2008visualizing} to project the single frame's 7-D vector (consisting of 7 metrics from \Tref{dataset_metrics}) to the corresponding 2D-coordinate for plotting the frame distribution of our dataset and the Chen21 dataset.  As shown in Fig.~\ref{fig:figure2}~(d), our dataset contains wider frame distribution than the Chen21 dataset, indicating that the networks trained with our dataset may be better generalized to different scenarios.  And the statistics of different datasets are shown in \Tref{7metrics}. In addition, our dataset contains more diverse motion patterns (see \Tref{simple comparison} and Fig.~\ref{fig:figure2}~(c)). The diversity in both scenes and motion patterns makes that our dataset can naturally be used for training deep networks and assessing the generalization capability of the networks across different scenes.
{\tiny }
\begin{table}[!t]
	\caption{Statistics of HDR labels in different datasets. Besides the DR, all numbers
		are in percentage.}
	\vspace{-0.4cm}
	\begin{center}
		\label{7metrics}
		\begin{threeparttable}
			\resizebox{\linewidth}{!}{
				\begin{tabular}{c|cc|ccc|cc}
					\toprule
					&\multicolumn{2}{c|}{\textbf{Extent of HDR}}&\multicolumn{3}{c|}{\textbf{Intra-frame Diversity}}&\multicolumn{2}{c}{\textbf{Overall-style}}\\
					\hline
					Dataset&\textbf{FHLP}&\textbf{EHL}&\textbf{SI}&\textbf{CF}&\textbf{stdL}&\textbf{ALL}&\textbf{DR}\\
					\hline
					Chen21~\cite{chen2021hdr}&8.85&2.46&8.93&2.77&11.35&5.29&2.54\\
					Ours&\textbf{13.75}&\textbf{2.72}&\textbf{9.16}&\textbf{4.30}&\textbf{12.13}&5.05&\textbf{2.73}\\
					\toprule
			\end{tabular}}
		\end{threeparttable}
		\vspace{-0.7cm}
	\end{center}
\end{table}

\begin{figure*}[!t]
	\centering
	\hspace{0.1cm}
	\includegraphics[width=6.5in]{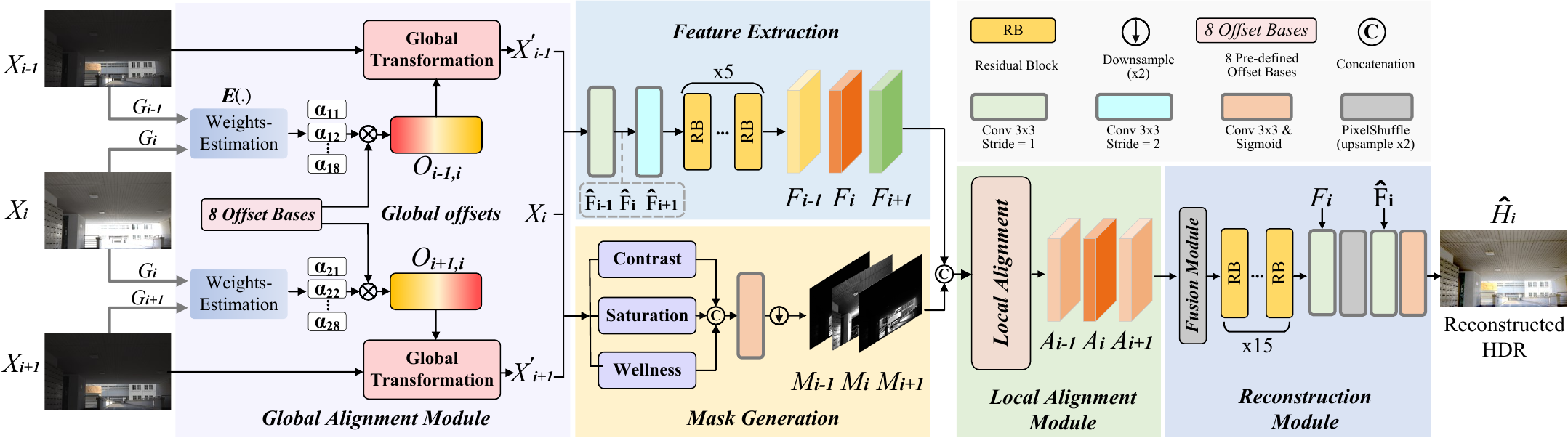}
	\caption{The architecture of our proposed network.}
	\label{backbone}
	\vspace{-0.45cm}
\end{figure*}

\section{Proposed Method}

Global motion (caused by camera movements) and local motion (caused by object motion) are almost inevitable when capturing videos, which imposes a core issue for HDR video reconstruction: how to perform alignment for the alternatingly-exposed inputs. Without effective alignment, the areas with motion in neighboring frames cannot be properly utilized to reconstruct the HDR frame, leading to severe ghosting artifacts. In this work, considering the differences between the global motion and local motion, we introduce a two-stage alignment network for HDR video reconstruction, which firstly performs alignment for the inputs (from global to local) and then adaptively fuses the aligned features to reconstruct the HDR video.

\paragraph{Overview}
Given an input LDR video $\lbrace I_{\mathit{i}}$$|$$\mathit{i}$ = 1,...,n$\rbrace$ with alternating exposures
$\lbrace t_{\mathit{i}}$$|$$\mathit{i}$ = 1,...,n$\rbrace$ \footnote{For example, the input sequences can alternate between two exposures $\lbrace$EV0, EV+3, EV0, EV+3, ...$\rbrace$ or three exposures $\lbrace$EV-2, EV0, EV+2, EV-2, EV0, EV+2, ...$\rbrace$.
In this work, we reconstruct the HDR video from sequences with two alternating exposures, while our network can be easily extended to other cases (\eg, three exposures).}, our target is to reconstruct the corresponding HDR video $\lbrace H_{\mathit{i}}$$|$$\mathit{i}$ = 1,...,n$\rbrace$. Following \cite{kalantari2019deep,chen2021hdr}, the input images are firstly mapped into the linear HDR domain by applying gamma correction:
\begin{equation}
	\label{inverse gamma}
	\bar{I_{\mathit{i}}} = I_{\mathit{i}}^{\gamma} \slash t_{\mathit{i}}, \hspace{0.1cm} \left(\gamma = 2.2\right).
\end{equation}
where $t_{\mathit{i}}$ is the exposure time of $I_{\mathit{i}}$. Then, the input image $I_{\mathit{i}}$ and the linear image $\bar{I_{\mathit{i}}}$ are concatenated into a 6-channels input $X_{\mathit{i}}$. Our network takes three continuous frames $\lbrace X_{\mathit{i-1}},X_{\mathit{i}},X_{\mathit{i+1}} \rbrace$ as input and predicts the HDR frame $\hat{H}_{i}$ for the center frame. As shown in Fig.~\ref{backbone}, our network consists of the global alignment module (GAM) for compensating global motion, the local alignment module (LAM) for compensating local motion, and the reconstruction module for reconstructing the HDR frame.

\paragraph{Global Alignment Module}
The global motion is relatively simple, which does not need to be modeled by dense pixel-wise optical flow with high Degree-of-Freedoms. Inspired by \cite{Ye2021, Wulff2015}, we use the pre-defined offset bases with 8 Degree-of-Freedoms (with each 2 for translation, rotation, scale, perspective~\cite{hartley2003multiple}) to model the global motion. Specifically, we design the GAM to estimate a weighted sum of 8 pre-defined offset bases for generating the global offsets. The global offsets are then used to spatially transform the inputs. Since all the operations in the GAM are differentiable, the GAM can be optimized through end-to-end training. In this way, the GAM can adaptively learn to compensate for the global motion between neighboring frames.

As shown in Fig.~\ref{backbone}, given the input $\lbrace X_{\mathit{j}}|\mathit{j} = \mathit{i}-1,\mathit{i},\mathit{i}+1\rbrace$, the GAM firstly uses a shared encoding layer to extract feature maps $G_{\mathit{j}}$ with 16 channels from inputs. Then, the features $\lbrace G_{\mathit{j}}|\mathit{j} = \mathit{i}-1,\mathit{i}+1\rbrace$ of neighboring frames are fed into the weights estimation module $E\left(.\right)$ (see our \textit{supplementary file} for the detailed architecture) along with the feature map $G_{\mathit{i}}$ of the reference image to obtain the corresponding weights $\lbrace\alpha_{1k}, \alpha_{2k}\rbrace$, generating the global offsets:
\vspace{-0.2cm}
\begin{subequations}
	\begin{equation}
		\label{offset estimates 1}
		O_{\mathit{i-1,i}} = \sum_{k=1}^{8} \mathit{\alpha}_{1k}\mathit{n}_{k} \quad \left(k = 1, 2, ..., 8\right),
	\end{equation}
        \vspace{-0.2cm}
	\begin{equation}
		\label{offset estimates 2}
		O_{\mathit{i+1,i}} = \sum_{k=1}^{8} \mathit{\alpha}_{2k}\mathit{n}_{k} \quad \left(k = 1, 2, ..., 8\right).
	\end{equation}
\end{subequations}
where the pre-defined offset bases $n_{\mathit{k}}$ are computed with the same settings in \cite{Ye2021}. The global offsets are then used to spatially transform the neighboring frames for compensating global motion between neighboring frames.

\paragraph{Local Alignment Module}
The LAM is designed to perform local alignment, which estimates the kernel weights at multiple scales in a coarse-to-fine manner and then performs a transformation for input features using adaptive separable convolution\cite{Niklaus2017} with the estimated kernel weights. In this way, the LAM can adaptively learn to integrate useful information in neighboring frames to compensate the missing content in reference frame, which facilitates the feature alignment under the noise and saturation. 

First, the shallow features $\lbrace F_{\mathit{j}}|\mathit{j} = \mathit{i}-1,\mathit{i},\mathit{i}+1\rbrace$ of input images are extracted. Inspired by Mertens~\etal~\cite{mertens2007exposure}, the contrast maps $c_{\mathit{j}}$, exposure wellness maps $e_{\mathit{j}}$, and saturation maps $s_{\mathit{j}}$ are extracted to provide the exposure information of the inputs. All three maps are concatenated together to generate the adaptive masks $M_{\mathit{j}}$ for input images.
\begin{figure}
	\centering
	\includegraphics[width=2.8in]{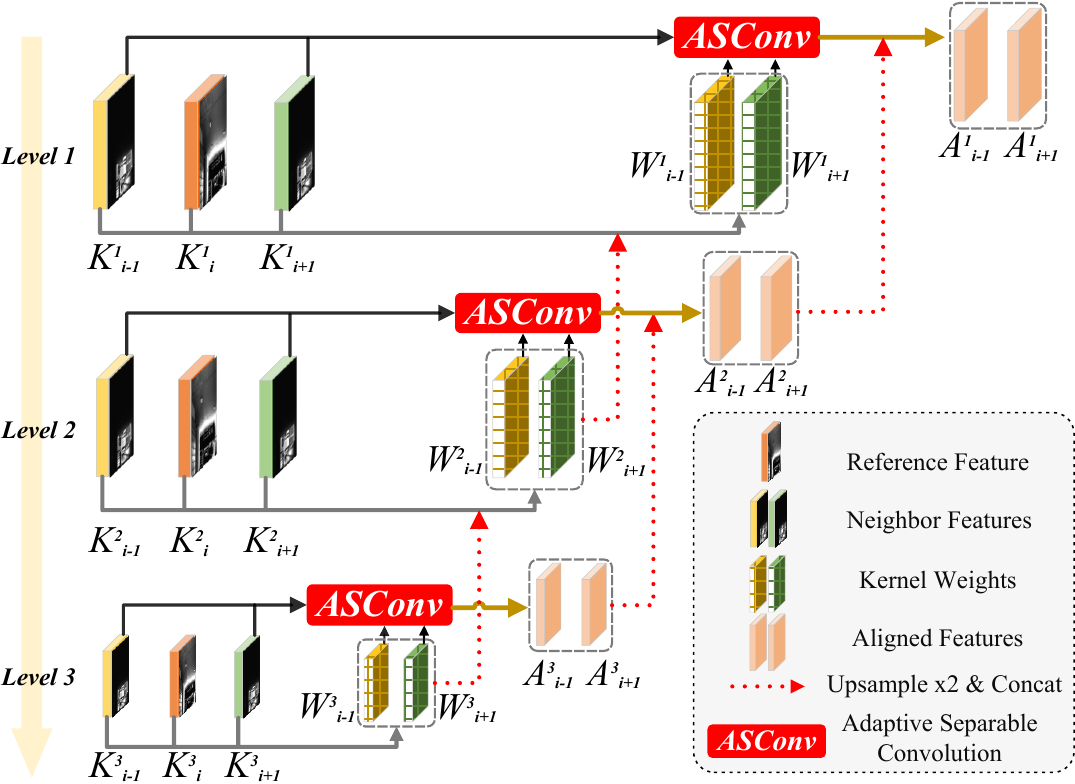}
	\caption{The architecture of local alignment module (LAM).}
	\label{LAM}
	\vspace{-0.25cm}
\end{figure} 
To handle large motions, the pyramidal processing is adopted, we generate an $L$-level pyramid of feature representation $\lbrace F^{l}_{\mathit{j}}|\mathit{j} = \mathit{i}-1,\mathit{i},\mathit{i}+1; \mathit{l} = 1,...,L\rbrace$ for input images and an $L$-level pyramid of masks $\lbrace M^{l}_{\mathit{j}}|\mathit{j} =  \mathit{i}-1,\mathit{i},\mathit{i}+1; \mathit{l} = 1,...,L\rbrace$. Then, we concatenate the corresponding masks and features along the channel dimension at each level to obtain the pyramid of tensors $\lbrace K^{l}_{\mathit{j}}|\mathit{j} = \mathit{i}-1,\mathit{i},\mathit{i}+1; \mathit{l} = 1,...,L\rbrace$, which are then utilized to predict kernel weights $W_{j}^{l}$ for neighboring features. With the predicted kernel weights, the aligned features $A^{l}_{\mathit{j}}$ can be obtained after performing adaptive separable convolution for neighboring features. 
Specifically, at the $l$-th level, kernel weights and aligned features are predicted also with the $\times$2 upsampled kernel weights and aligned features from the upper $\left(l+1\right)$-th level, respectively (red dot lines in Fig.~\ref{LAM}):
\vspace{-0.2cm}
\begin{subequations}
	\begin{equation}
		\label{upsampe1}
		\begin{aligned}
			&W_{i-1}^{l},W_{i+1}^{l} = g\Big(\lbrack K^{l}_{\mathit{i-1}},K^{l}_{\mathit{i}},K^{l}_{\mathit{i+1}},\\
			&\left(W_{i-1}^{l+1}\right)^{\uparrow\times2},\left(W_{i+1}^{l+1}\right)^{\uparrow\times2}\rbrack \Big),\\
		\end{aligned}
	\end{equation}
        \vspace{-0.5cm}
	\begin{equation}
		\label{seprable conv 1}
		A_{i-1}^{l} = h\left(ASConv\left(F^{l}_{\mathit{i-1}},W_{i-1}^{l}\right),\left(A_{i-1}^{l+1}\right)^{\uparrow\times2}\right),
	\end{equation}
        \vspace{-0.5cm}
	\begin{equation}
		\label{seprable conv 2}
		A_{i+1}^{l} = h\left(ASConv\left(F^{l}_{\mathit{i+1}},W_{i+1}^{l}\right),\left(A_{i+1}^{l+1}\right)^{\uparrow\times2}\right).
	\end{equation}
        \vspace{-0.3cm}
\end{subequations}

where $\left(.\right)^{\uparrow\times2}$ is the upscaling operation with a factor of 2, $\lbrack . \rbrack$ is the concat operation, $g\left(.\right)$ is the kernel weights predictor consisting of several convolution layers, $ASConv\left(.\right)$ denotes the adaptive separable convolution, $h\left(.\right)$ is the general function with several convolution layers. We use three-level pyramid, \ie, L=3, in LAM. The kernel size is set to 31 in the adaptive separable convolution.

\begin{figure}
 	\centering
 	\includegraphics[width=2.9in]{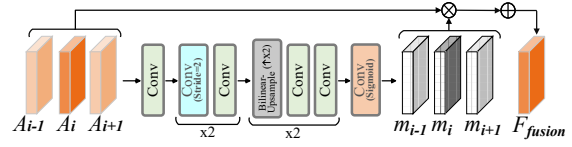}
 	\caption{The architecture of fusion module.}
 	\label{fusenet}
 	\vspace{-0.2cm}
 \end{figure}

\paragraph{Fusion and Reconstruction}
The fusion module is used to fuse the aligned features, which can suppress the harmful features from under-exposed and over-exposed areas. As shown in Fig.~\ref{fusenet}, the aligned features are concatenated as the input of the fusion module, generating fusion masks for fusing the aligned features $A_{j}$. 
Then, the fused feature $F_{fusion}$ is passed through a series of residual blocks. Two skip connections are added to concatenate shallow features of the reference frame. Finally, the HDR frame $\hat{H}$ can be obtained after a convolution layer and a sigmoid activation layer.

\paragraph{Loss Function}
Since HDR images are typically displayed after tonemapping, following~\cite{kalantari2019deep, chen2021hdr,yan2019attention}, we use the $\mu$-law tonemapping function to the HDR image:
\begin{equation}
	\label{ufunction}
	T \left ( H \right ) = \frac{log \left ( 1 + \mu H \right )}{log \left ( 1 + \mu \right )}, \mu = 5000.
\end{equation} 
where $T \left ( H \right )$ denotes the tonemapped HDR image. We train the network by minimizing the $l$1 loss distance $L = || T \big(\hat{H}\big) - T\big( H \big) ||$ between the tonemapped estimated $\hat{H}$ and ground truth $H$.

\section{Experiments}
\subsection{Experimental Settings}
\paragraph{Datasets}
There are three datasets adopted, including our Real-HDRV, the Chen21 dataset~\cite{chen2021hdr} and the synthetic dataset~\cite{chen2021hdr}. As for our dataset, it is divided into the training collection (450 videos) and the testing collection (27 videos for indoor daytime and outdoor daytime scenes, 23 videos for indoor nighttime and outdoor nighttime scenes). 
Each video in the testing collection provides 8 LDR frames with two alternating exposures and the corresponding HDR labels.
As for the synthetic dataset, we utilized 21 existing HDR videos from ~\cite{froehlich2014creating, Kronander2014} and Vimeo-90K~\cite{xue2019video} to synthesize the dataset with the same settings as in~\cite{chen2021hdr}.
The Chen21 dataset contains 76 dynamic image pairs, 49 static image pairs augmented with random global motion, and 50 unlabeled sequences with two alternating exposures.

\renewcommand\arraystretch{1.2}
\begin{table*}[!t] 
	\centering
	\setlength{\abovecaptionskip}{-0.3cm} 
	\setlength{\belowcaptionskip}{-0.2cm}
	\caption{ Quantitative comparison for training on the synthetic dataset~\cite{chen2021hdr} or our dataset, while evaluating on the Chen21 dataset~\cite{chen2021hdr}. The better results are highlighted in bold. Among these evaluation metrics, the higher quality of the tested HDR image leads to the higher score.}
	\begin{center}
		\resizebox{\textwidth}{!}{
			\Large
			\begin{tabular}{c||c||cccccc||cccccc}
				\toprule[1.4pt]
				& Training &\multicolumn{6}{c||}{Evaluation on the dynamic set} & \multicolumn{6}{c}{Evaluation on the static set} \\
				Methods & dataset & PSNR-$\mu$ & SSIM-$\mu$ & PU-PSNR &PU-SSIM & HDR-VDP-2 & HDR-VQM & PSNR-$\mu$ & SSIM-$\mu$ & PU-PSNR & PU-SSIM & HDR-VDP-2 & HDR-VQM  \\
				\midrule    
				\multirow{2}{*}{AHDRNet~\cite{yan2019attention}} 
				& Synthetic & 44.34 & 0.9668 & 38.48 & 0.9718 & 62.05 & 84.56 & 38.38 & 0.9329 & 32.99 & 0.9422 & 58.19 & 71.40 \\
				& Our & \textbf{45.02} & \textbf{0.9741} & \textbf{39.17} & \textbf{0.9808} & \textbf{62.51} & \textbf{89.60} & \textbf{40.16} & \textbf{0.9589} & \textbf{34.69} & \textbf{0.9638} & \textbf{59.53} & \textbf{77.84} \\    		                                                                                                              
				\multirow{2}{*}{Kalantari19~\cite{kalantari2019deep}} 
				& Synthetic & 44.15 & 0.9637 & 38.39 & 0.9728 & 59.44 & 86.56 & 40.59 & 0.9316 & 35.22 & 0.9429 & 57.53 & 74.60 \\
				& Our & \textbf{45.31} & \textbf{0.9689} & \textbf{39.37} & \textbf{0.9757} & \textbf{61.39} & \textbf{86.95} & \textbf{41.19} & \textbf{0.9336} & \textbf{36.03} &0.9429 & \textbf{59.69} & \textbf{81.83} \\  
				\multirow{2}{*}{Chen21~\cite{chen2021hdr}} 
				& Synthetic & 45.46 & 0.9706 & 39.46 & 0.9760 & 61.26 & 87.40 & 41.21 & 0.9412 & 35.81 & 0.9483 & 59.19 & 78.98 \\
				& Our & \textbf{45.65} & \textbf{0.9716} & \textbf{39.79} & \textbf{0.9768} & \textbf{61.39} & \textbf{90.33} & \textbf{41.37} & \textbf{0.9419} &\textbf{36.20} & \textbf{0.9516} & \textbf{59.46} & \textbf{81.43} \\  
				\multirow{2}{*}{CA-ViT~\cite{liu2022ghost}} 
				& Synthetic & 44.76 & 0.9664 & 38.81 & 0.9714 & 61.95 & 88.78 & 38.26 & 0.9252 & 32.84 & 0.9368 & 58.27  & 71.90 \\
				& Our & \textbf{45.19} & \textbf{0.9744} & \textbf{39.33} & \textbf{0.9814} & \textbf{62.43} & \textbf{90.41} & \textbf{39.91} & \textbf{0.9570} & \textbf{34.30} & \textbf{0.9609} & \textbf{59.05}  & \textbf{77.69} \\  
				\multirow{2}{*}{LAN-HDR~\cite{chung2023lan}} 
				& Synthetic & 44.81 & 0.9714 & 38.64 & 0.9773 & 61.64 & 88.86 & 39.34 & 0.9424 & 33.87 & 0.9490 & 57.12 & 70.47\\
				& Our & \textbf{45.38} & \textbf{0.9743} & \textbf{39.53} &\textbf{ 0.9804} & \textbf{62.83} & \textbf{88.96}  & \textbf{40.09} & \textbf{0.9565} & \textbf{34.63} & \textbf{0.9595} & \textbf{59.78} & \textbf{79.29} \\  
				\toprule[1.4pt]
		\end{tabular}}
	\end{center}
	\label{comparison dataset}
	\vspace{-0.6cm}
\end{table*}


\paragraph{Implementation Details}
We generate LDR sequences with two alternate exposures separated by three stops for each video in our training collection. We then sample three LDR frames as input and produce the HDR label for the center frame to generate a training sample.
We crop patches of size 256$\times$256 with a stride of 128 from the training set for training. Random rotation and flipping augmentation are applied. 
We use Adam optimizer, and set the batch size and initial learning rate as 4 and 0.0001, respectively. We implement our model using PyTorch with 6 NVIDIA 3090 GPUs and train for 100 epochs.

\paragraph{Evaluation Metrics}
 We use six common metrics for testing, \ie, HDR-VDP-2~\cite{mantiuk2011hdr}, PSNR-$\mu$, SSIM-$\mu$, PU-PSNR, PU-SSIM and HDR-VQM~\cite{narwaria2015hdr}. PSNR-$\mu$ and SSIM-$\mu$ are computed after tonemapping with $\mu$-law function (in \Eref{ufunction}). PU-PSNR and PU-SSIM are computed after perceptually uniform encoding~\cite{Mantiuk2021}. When computing the HDR-VDP-2, the diagonal display size is set to 30 inches.

\begin{figure}[t] 
	\centering 
	\hspace{0.35cm}
	\includegraphics[width=0.45\textwidth]{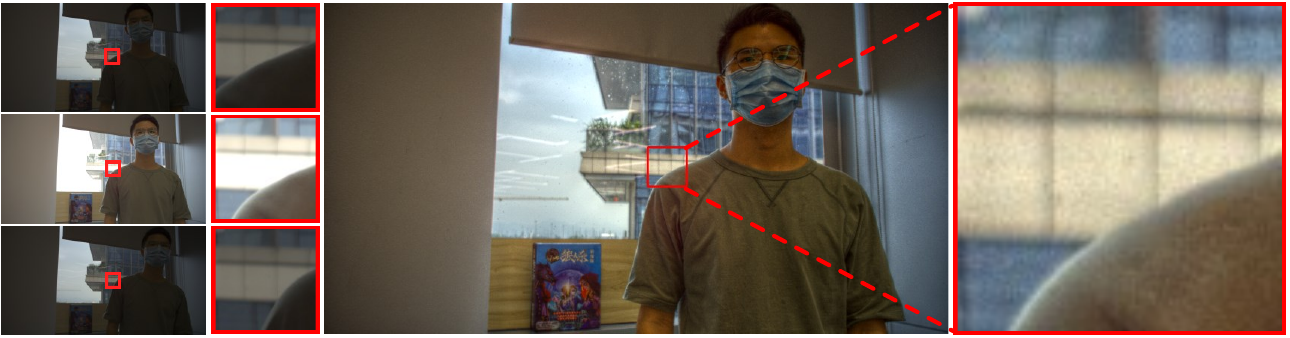}
	\\
	\vspace{-0.19cm}
	\makebox[0.06\textwidth]{\scriptsize \hspace{-2.2cm} LDR inputs }
	\makebox[0.12\textwidth]{\scriptsize  \hspace{0.5cm}  HDR label }
	\makebox[0.06\textwidth]{\scriptsize \hspace{2.7cm} HDR patch }
	\\
	\vspace{+0.03cm}
	\hspace{-0.2cm}
	\rotatebox{90}{\scriptsize \hspace{1.6em}\textbf{\makecell{Our\\dataset}} \hspace{1.8em} \makecell{Synthetic\\dataset}}
	\hspace{-0.2cm}
	\includegraphics[width=0.45\textwidth]{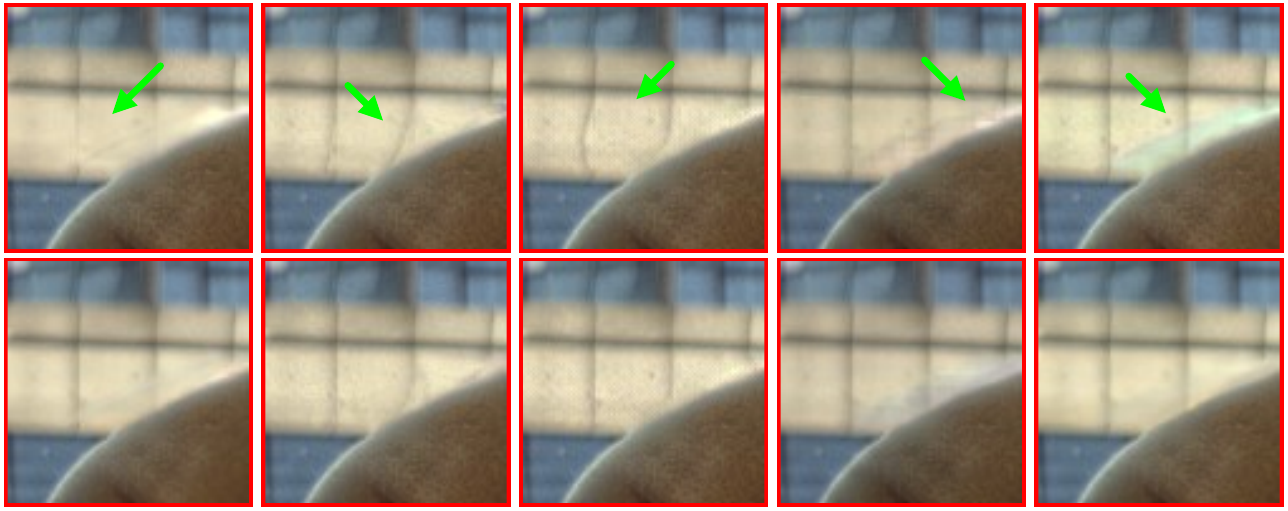}
	\\
	\vspace{-0.17cm}
	\hspace{0.1cm}
	\makebox[0.09\textwidth]{\scriptsize \hspace{0.05cm} AHDRNet\cite{yan2019attention}}
	\makebox[0.09\textwidth]{\scriptsize \hspace{0.05cm} Kalantari19\cite{kalantari2019deep}}
	\makebox[0.08\textwidth]{\scriptsize  \hspace{0.16cm} Chen21\cite{chen2021hdr}}
	\makebox[0.08\textwidth]{\scriptsize \hspace{0.20cm} CA-ViT\cite{liu2022ghost}}
	\makebox[0.09\textwidth]{\scriptsize \hspace{0.25cm} LAN-HDR\cite{chung2023lan}}
	\\
	\caption{ Visual comparison of different models trained on the synthetic dataset~\cite{chen2021hdr} and our dataset. } 
	\label{dataset compare 1}
	\vspace{-0.35cm}
\end{figure}


\subsection{Evaluation of Our Proposed Dataset}
To evaluate the effectiveness of our dataset, we compare our dataset with the synthetic dataset~\cite{chen2021hdr}. We train representative HDR reconstruction models~\cite{yan2019attention, kalantari2019deep,chen2021hdr,liu2022ghost,chung2023lan} on our dataset and the synthetic dataset, and evaluate the performance of trained models on the Chen21 dataset~\cite{chen2021hdr}.

\paragraph{Quantitative Results} 
As shown in \Tref{comparison dataset}, the models trained on our dataset can achieve better performance on the real-world dataset~\cite{chen2021hdr} than the models trained on the synthetic dataset, demonstrating the effectiveness of our dataset. For example, compared with Kalantari19 trained on the synthetic dataset, the same model trained on our dataset can achieve more than 1 dB gain (PSNR-$\mu$) on the dynamic set of Chen21 dataset, which is significant. Similar improvements can also be observed in other methods. 

\paragraph{Qualitative Results} 
The visual comparison for the models trained on different datasets is provided in Fig.~\ref{dataset compare 1}. Obviously, the models trained on our dataset yield better visual quality, while the models trained on the synthetic dataset typically yield severe ghosting artifacts or color distortions.
The superior performance of the models trained with our datasets comes from the real degradation distribution in our dataset (more qualitative comparisons are provided in \textit{supplementary file}). In summary, models trained on our Real-HDRV can better handle real-world scenes, demonstrating the effectiveness of our dataset.
\renewcommand\arraystretch{1.2}
\begin{table*}
	\setlength{\abovecaptionskip}{-0.3cm} 
	\setlength{\belowcaptionskip}{-0.15cm}
	\caption{Quantitative comparison of our method with state-of-the-art methods on our dataset. \textcolor{red}{Red} text indicates the best and \textcolor{blue}{blue} text indicates the second best result, respectively. ID\&OD denotes indoor daytime and outdoor daytime scenes. IN\&ON denotes indoor nighttime and outdoor nighttime scenes. }
	\begin{center}
		\label{result_on_Ours}
		\resizebox{\textwidth}{!}{
			\begin{threeparttable}
				\begin{tabular}{c||ccc||ccc||ccc||ccc||ccc||ccc}
					\toprule[1.4pt]
					&\multicolumn{3}{c||}{PSNR-$\mu$}&\multicolumn{3}{c||}{SSIM-$\mu$}&\multicolumn{3}{c||}{PU-PSNR}&\multicolumn{3}{c||}{PU-SSIM}&\multicolumn{3}{c||}{HDR-VDP-2}&\multicolumn{3}{c}{HDR-VQM}\\
					Methods&ID\&OD&IN\&ON&Avg&ID\&OD&IN\&ON&Avg&ID\&OD&IN\&ON&Avg&ID\&OD&IN\&ON&Avg&ID\&OD&IN\&ON&Avg&ID\&OD&IN\&ON&Avg\\
					\cline{1-19}
					Kalantari13\cite{kalantari2013patch}& 41.27&37.24 &39.41 & 0.9697&0.9246 &0.9490 &35.11 &31.76 &33.57 & 0.9738& 0.9439& 0.9601&58.03 &56.42 &57.30 & 90.67&78.80&85.21\\
					AHDRNet\cite{yan2019attention}&45.02 &40.72 &43.06 & 0.9840&0.9613 & 0.9736& 38.96& 35.08& 37.17&0.9831 &0.9666 & 0.9755&60.53 &57.78 &59.27 & 90.20&77.60&84.40\\
					Kalantari19\cite{kalantari2019deep}& 43.90& 39.23&41.75 & 0.9769&0.9437 &0.9616 &38.07 &33.99 & 36.19&0.9797 &0.9576 & 0.9695& 61.61& 59.05&60.43 & 92.13 &82.16 &87.54\\
					Chen21\cite{chen2021hdr}&44.42 & 39.43& 42.12&0.9789 &0.9466 &0.9640 &38.70 & 34.29& 36.67& 0.9832&0.9612 &0.9731&\textcolor{blue}{63.89}&\textcolor{blue}{60.70}&\textcolor{blue}{62.41} & \textcolor{blue}{94.22}&\textcolor{blue}{83.93}&\textcolor{blue}{89.49}\\
					CA-ViT~\cite{liu2022ghost}& 44.65&40.16 &42.58 &0.9834 &0.9603 &0.9728 &38.44 & 34.34 &36.55 &0.9820 & 0.9651&0.9743 &60.62 &57.38 & 59.13& 90.38&77.63&84.51 \\
					Yue23 \cite{yue2023hdr}& \textcolor{blue}{45.42}&\textcolor{blue}{41.20}& \textcolor{blue}{43.48}&\textcolor{blue}{0.9849}&\textcolor{blue}{0.9626}&\textcolor{blue}{0.9746} &\textcolor{blue}{39.37}&\textcolor{blue}{35.65}&\textcolor{blue}{37.66} & \textcolor{blue}{0.9845}&\textcolor{blue}{0.9682}&\textcolor{blue}{0.9770}&62.93&59.70&61.44& 93.23&82.99&88.52\\
					LAN-HDR\cite{chung2023lan}& 44.23& 40.54 &42.53 &0.9836 &0.9607 &0.9731 &38.16 &34.93 &36.67 &0.9823 & 0.9646&0.9742 &60.89 &58.38 &59.73 & 91.28&78.73&85.51 \\
					Ours&\textcolor{red}{45.81}&\textcolor{red}{41.58}&\textcolor{red}{43.86} &\textcolor{red}{0.9856}&\textcolor{red}{0.9643}&\textcolor{red}{0.9758} &\textcolor{red}{39.83}&\textcolor{red}{36.10}&\textcolor{red}{38.11} &\textcolor{red}{0.9860}&\textcolor{red}{0.9711}&\textcolor{red}{0.9792} &\textcolor{red}{65.34}&\textcolor{red}{61.40}&\textcolor{red}{63.53}& \textcolor{red}{94.47}&\textcolor{red}{84.13}&\textcolor{red}{89.71}\\
					\cline{1-9}
					\toprule[1.4pt]
				\end{tabular}
		\end{threeparttable}}
		\vspace{-0.47cm}
	\end{center}
\end{table*}

\begin{figure}[t] 
	\centering 
	\hspace{0.35cm}
	\includegraphics[width=0.45\textwidth]{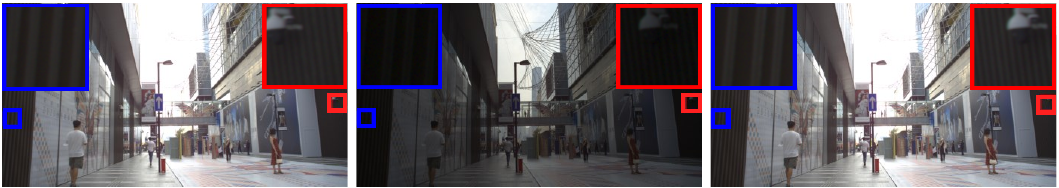}
	\\
	\vspace{-0.19cm}
	\makebox[0.108\textwidth]{\scriptsize \hspace{1.1em} Alternatingly-exposed LDR inputs (\textit{Middle frame} is the reference frame)}
	\\
	\vspace{+0.03cm}
	\hspace{-0.2cm}
	\rotatebox{90}{\scriptsize \hspace{0.9em}\textbf{\makecell{Our\\dataset}} \hspace{1.8em} \makecell{Synthetic\\dataset}}
	\hspace{-0.2cm}
	\includegraphics[width=0.45\textwidth]{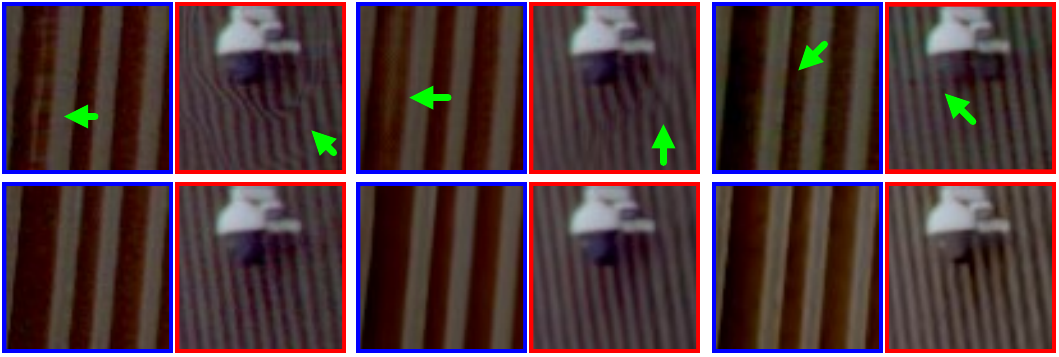}
	\\
	\vspace{-0.17cm}
	\hspace{0.15cm}
	\makebox[0.14\textwidth]{\scriptsize Kalantari19\cite{kalantari2019deep}}
	\makebox[0.14\textwidth]{\scriptsize Chen21\cite{chen2021hdr}}
	\makebox[0.14\textwidth]{\scriptsize LAN-HDR\cite{chung2023lan}}
	\\
	\caption{Visual example of models trained on different datasets. } 
	\label{dataset compare 2 }
	\vspace{-0.35cm}
\end{figure}
\begin{figure*}[!t]
	\centering
	\includegraphics[width=6.8in]{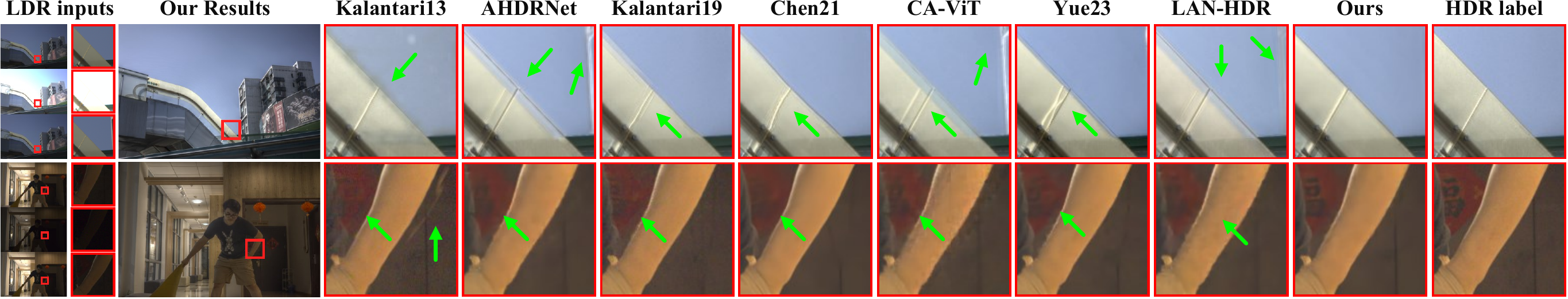}
	\caption{Visual comparison of different networks trained on our dataset. Please zoom in for more details. }
	\label{compare_on_our}
	\vspace{-0.25cm}
\end{figure*}
\paragraph{Evaluation on Unlabeled Real-World Dataset} We also evaluate the varying models on unlabeled sequences of the Chen21 dataset~\cite{chen2021hdr}. The visual comparison is provided in Fig.~\ref{dataset compare 2 }. As seen, when the reference frame is low-exposure, the models trained on our dataset can recover clear details, while the models trained on the synthetic dataset generate corrupted details or color distortions. Similar improvements can be observed when the reference frame is high-exposure, please refer to our \textit{supplementary file} for more details.

\subsection{Evaluation of Our Proposed Method}
We compare our method with prevalent state-of-the-art HDR video reconstruction methods~\cite{kalantari2013patch, kalantari2019deep, chen2021hdr, yue2023hdr, chung2023lan} and state-of-the-art HDR deghosting methods~\cite{yan2019attention, liu2022ghost} on our dataset for a comprehensive evaluation. For a fair comparison, we use their officially released codes, if accessible, otherwise, we re-implement their methods based on their papers. Note that the AHDRNet~\cite{yan2019attention} and CA-ViT~\cite{liu2022ghost} are adapted for HDR video reconstruction by changing the network input. In addition, we evaluate our method on the Chen21 dataset\cite{chen2021hdr} to demonstrate the generalization of our method (more details can be found in \textit{supplementary file}).

\paragraph{Quantitative Results} 
The quantitative comparison between our method and other methods is listed in \Tref{result_on_Ours}. Compared to other methods, our method achieves the best average performance in all the evaluation metrics, demonstrating the effectiveness of our method. In addition, evaluated in different scenes, our method can also acquire the best performance, demonstrating that our method can better handle sequences under different real scenes. 

\paragraph{Qualitative Results} 
The visual comparison of varying methods on our dataset is shown in Fig.~\ref{compare_on_our}. Obviously, our method achieves more excellent visual quality than other methods, which can recover the missing content of the over-exposed areas without introducing artifacts when the reference frame is high-exposure (see the $1^{st}$ row in Fig.~\ref{compare_on_our}). Also, our method can better remove the noise and faithfully preserve the structure of the under-exposed areas when the reference frame is low-exposure (see the $2^{nd}$ row in Fig.~\ref{compare_on_our}). 
In contrast, due to the inaccurate-prone flow, the flow-based methods \cite{kalantari2013patch, kalantari2019deep, chen2021hdr,yue2023hdr} usually suffer from unpleasing artifacts for the over-exposed areas, and they cannot faithfully recover the details in the under-exposed areas. 
Additionally, due to the lack of effective alignment, the attention-based methods~\cite{yan2019attention, liu2022ghost} can easily introduce ghosting artifacts.

\begin{table}[!t]
	\caption{Computation complexity comparison. All methods are executed on a Nvidia 3090 GPU.}
	\vspace{-0.45cm}
	\begin{center}
			\label{complexity}
			\resizebox{\linewidth}{!}{
				\begin{threeparttable}
						\begin{tabular}{c|ccccccc}
								\toprule
								Methods&Kalantari19\cite{kalantari2019deep}&Chen21\cite{chen2021hdr}&Yue23\cite{yue2023hdr}&LAN-HDR\cite{chung2023lan}&Ours\\
								\hline
								Params. (M) & 10.39 & 6.44 & 3.5 & 7.31 & 5.98\\
								Flops (T) & 2.13 & 5.29 & 3.96 & 1.24 & 2.07 \\
								Time (s) & 0.24 & 0.87 & 0.76 & 0.55 & 0.34\\
								\toprule
							\end{tabular}
					\end{threeparttable}}
		\end{center}
	\vspace{-0.6cm}
\end{table}

\paragraph{Complexity Comparisons} 
For each test method, we record the quantity of the model parameters, the execution time and the floating point operations (flops) of generating an HDR frame with the size of 1500 $\times$ 1000. As shown in~\Tref{complexity}, our model can achieve the best trade-off between the performance and the computational cost.
\renewcommand\arraystretch{1.2}
\begin{table}
	\caption{Quantitative results of the ablation studies. Our baseline network uses the same architecture as our full model (Model4), but with the GAM and the LAM removed}
	\vspace{-0.45cm}
	\begin{center}
		\label{Ablation}
		\resizebox{\linewidth}{!}{
		\begin{threeparttable}
			\begin{tabular}{c|ccc|ccc}
				\toprule
				Models&Baseline&GAM&LAM&HDR-VDP-2&PSNR-$\mu$&HDR-VQM\\
				\hline
				Model1&$\checkmark$&\scalebox{0.75}[1]{$\times$}&\scalebox{0.75}[1]{$\times$}&59.49&42.67&85.34\\
				Model2&$\checkmark$&$\checkmark$&\scalebox{0.75}[1]{$\times$}&61.72&43.07&87.62\\
				Model3&$\checkmark$&\scalebox{0.75}[1]{$\times$}&$\checkmark$&62.58&43.61&88.45\\
				Model4&$\checkmark$&$\checkmark$&$\checkmark$&63.53&43.86&89.71\\
				\toprule
			\end{tabular}
		\end{threeparttable}}
	\end{center}
	\vspace{-0.6cm}
\end{table}

\paragraph{Ablation Study}
To analyze the effectiveness of each component in our network, we conduct comprehensive ablation studies on our dataset. As shown in \Tref{Ablation}, the GAM and LAM both improve the performance, demonstrating the effectiveness of the GAM and the LAM. On the one hand, the model (Baseline) performs poorly when directly conducting HDR video reconstruction without performing alignment, demonstrating that the alignment is very critical to HDR video reconstruction. On the other hand, by using the GAM to perform global alignment, our full model can more effectively handle the complex motion, obtaining the higher HDR-VDP-2 score and the higher HDR-VQM score than ours (w/o GAM). Also, our full model can achieve better performance than ours (w/o LAM). 

\section{Conclusion}
We constructed a novel dataset for HDR video reconstruction, which contains various scenes, diverse motion patterns, and high-quality labels. Our dataset can also be applied to other HDR tasks for future research.
Then, we proposed a novel framework for HDR video reconstruction, which considers the differences between global motion and local motion. The designed GAM enables our framework to better handle global motion. And the designed LAM can adaptively integrate the useful information in neighboring frames to help reconstruct the reference HDR frame, effectively decreasing the ghosting artifacts caused by large local motion. Extensive experiments demonstrate the superiority of our dataset and our method. 
{
    \small
    \bibliographystyle{ieeenat_fullname}
    \bibliography{main}
}


\end{document}